\documentclass[lettersize,journal]{IEEEtran}
\usepackage{amsmath,amsfonts}
\usepackage{algorithmic}
\usepackage{algorithm}
\usepackage{array}
\usepackage[caption=false,font=normalsize,labelfont=sf,textfont=sf]{subfig}
\usepackage{textcomp}
\usepackage{stfloats}
\usepackage{url}
\usepackage{verbatim}
\usepackage{graphicx}
\usepackage{cite}
\usepackage{xcolor}
\usepackage[hidelinks]{hyperref}
\begin{document}
\title{Neuromorphic quadratic programming for efficient and scalable model predictive control}

\author{Ashish Rao Mangalore*,
        Gabriel Andres Fonseca Guerra,
        Sumedh R. Risbud,
        Philipp Stratmann, Andreas Wild
        \thanks{* Corresponding author \\ 
        A. R. Mangalore is presently with the School of Computation, Information, and Technology, Technische Universit\"at M\"unchen and Neuromorphic Computing Lab, Intel Labs. G. A. F. Fonseca, S. R. Risbud, P. Stratmann, and A. Wild are with the Neuromorphic Computing Lab, Intel Labs. E-mail: \{ashish.rao.mangalore, gabriel.fonseca.guerra, sumedh.risbud, philipp.stratmann, andreas.wild\}@intel.com}
        }

\markboth{Accepted by Robotics and Automation Magazine}%
{Shell \MakeLowercase{\textit{et al.}}: A Sample Article Using IEEEtran.cls for IEEE Journals}


\maketitle

\begin{abstract}
  Applications in robotics or other size-, weight- and power-constrained autonomous systems at the edge often require real-time and low-energy solutions to large optimization problems. 
Event-based and memory-integrated neuromorphic architectures promise to solve such optimization problems with superior energy efficiency and performance compared to conventional von Neumann architectures. 
Here, we present a method to solve convex continuous optimization problems with quadratic cost functions and linear constraints on Intel’s scalable neuromorphic research chip \textit{Loihi 2}. When applied to model predictive control (MPC) problems for the quadruped robotic platform \textit{ANYmal}, this method achieves over two orders of magnitude reduction in combined energy-delay product compared to the state-of-the-art solver, OSQP, on (edge) CPUs and GPUs with solution times under ten milliseconds for various problem sizes. These results demonstrate the benefit of non-von-Neumann architectures for robotic control applications.

\end{abstract}


\section{Introduction}
\label{sec:intro}
\begin{figure*}
    \centering
    \includegraphics[width=0.75\textwidth]{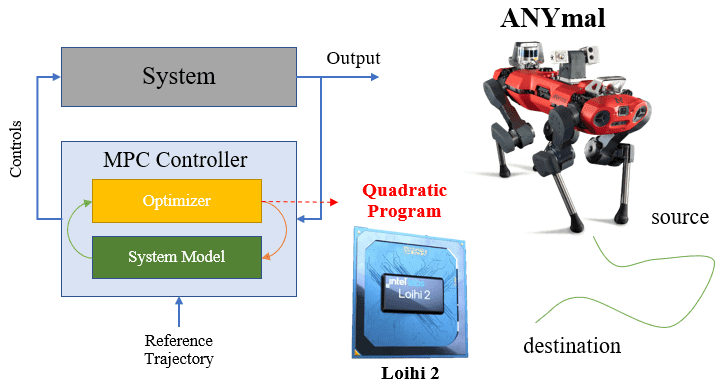}
    \caption{A class of convex optimization problems, namely, quadratic programs, are part of the model predictive control loop of the ANYmal. In this article, we explore running different sizes of these programs on Loihi 2, a neuromorphic research platform developed by Intel Labs and see how they compare against conventional compute architectures in terms of performance and energy consumption. We show how the common motifs underlying many algorithms (such as gradient descent, primal-dual or operator splitting methods) to solve convex constrained optimization problems are framed as the dynamics of event-based recurrent neural networks whose steady-states represent the solutions to the problems.
    The resulting network topology enables the implementation of diverse first-order algorithms for efficiently solving convex QP and LP problems on neuromorphic hardware.}
    \label{fig:visual_abstract}
\end{figure*}
\IEEEPARstart{C}{onvex} quadratic programming (QP) has been a topic of substantial research since the 1950s. The goal of this class of problems is to optimize a quadratic cost function subject to linear constraints. The convex nature of the problems ensures that iterative updates of the variables along the gradient of the cost function are guaranteed to converge to the optimal solution. Convex QP optimization problems are particularly attractive for edge applications like robotic model predictive control (MPC), in which a smooth closed-loop interaction with the environment requires solving the problems within millisecond latency \cite{sleiman2021unified}. In such embedded applications, energy consumption is also critical for long battery life. As control systems incorporate more degrees of freedom, the underlying optimization problems grow in terms of the number of variables and require more complex cost and constraint functions. Solving increasingly more difficult optimization problems drives the search for more efficient and scalable approaches beyond conventional CPUs or GPUs.

Brain-inspired neuromorphic architectures have demonstrated significant performance and energy gains over conventional architectures for a range of optimization problems with superior scalability up to hundred of thousands of problem variables \cite{schuman2022}.
Neuromorphic architectures derive their advantage over conventional  architectures from the integration of memory with compute units to minimize data movement, massive fine-grained parallelism, a streamlined set of supported operations, as well as architectural optimizations enabling sparse, event-based computation and communication only when necessary.
As a result and similar to biological brains, these novel architectures have the potential to solve extremely complicated computational problems at low power and short response times on the order of Watt and millisecond, respectively.

Among the algorithms that excel on neuromorphic architectures are solvers for constraint satisfaction problems \cite{fonseca2017CSP}, 
quadratic unconstrained binary optimization (QUBO) \cite{alom2017}, and different optimization problems on graphs \cite{mniszewski2019}. The development of the spiking locally competitive algorithm, LCA, \cite{tang2017spikingLCA, loihi2021survey} \textemdash{} to solve LASSO with neuromorphic hardware \textemdash{} was the first approach to solve unconstrained convex QPs as a spiking neural network with wide ranging applications such as in sparse coding or signal processing. Nonetheless, the problem of solving general convex optimization with constraints on neuromorphic hardware remained unaddressed.

In this article, we discuss a framework to solve general convex QPs on neuromorphic hardware and demonstrate the implementation of a QP solver that leverages the event-based, memory-integrated, fine-granular parallel architecture of the Intel \textit{Loihi 2} research chip \cite{orchard2021} using the Lava open-source framework\footnote{\href{https://lava-nc.org/}{https://lava-nc.org/}}. We further, highlight its efficacy to solve large real-world QP problems arising in the context of model predictive control (MPC) of the ANYmal quadrupedal robot \cite{hutter2016anymal}. We explore the conditions under which neuromorphic architectures are more suitable hardware substrates to solve convex optimization problems than traditional von-Neumann-based architectures.

Convex QP problems arising in size-, weight-, and power-constrained (SWaP) systems are conventionally solved on CPUs, even though a few solutions have been developed to leverage the parallel compute capabilities of GPUs, FPGAs, or ASICs. A range of high-performance QP solvers exist for CPUs, such as \textit{GUROBI} \cite{gurobi}, \textit{MOSEK} \cite{mosek}, SCS \cite{odonoghue:21}, and \textit{CVXOPT} \cite{cvxopt}. 
Lightweight CPU solvers specifically optimized for embedded systems include \textit{qpOASES} \cite{Ferreau2014}, \textit{ECOS} \cite{domahidi2013} and \textit{OSQP} \cite{stellato2020}; such solvers avoid any dependence on large external libraries, use only basic operations (e.g. avoiding division), minimize the steps to solution, optimize the code for mobile processors, or often parallelize their code. Unfortunately, CPUs in general \textemdash{} and the ones in embedded systems in particular \textemdash{} often do not support the degree of parallelism needed to accelerate large optimization workloads.

While GPUs offer a high degree of parallelism for optimization algorithms \cite{yu2017, schubiger2020}, they primarily achieve their efficiency through extremely wide data paths and deep pipe-lining, allowing them to stream batched data from off-chip memory to process many parallel threads. However, in the case of sparse problems, the GPU resources are massively underutilized, rendering them inefficient.  Similar to sparse problems, the inefficiencies of using GPUs become apparent in real-time data processing applications too (e.g., MPC), wherein in the absence of batching (or batch-size 1) leads to inadequate usage of pipe-lining.
In addition, both CPUs and GPUs suffer from high external memory access latencies that can hardly be hidden when solving iterative algorithms on a millisecond timescale in closed-loop control.

Similar to neuromorphic processors, more specialized solutions for solving QPs have been realized using FPGAs \cite{mkinerney2018,lucia2018} and ASICs \cite{skibik2018}. While these approaches have their merits in terms of performance or energy consumption, they require a vastly higher development effort and may only serve a single purpose (ASICs) compared to highly efficient and programmable neuromorphic processors like Intel Loihi that can be applied to many other problems apart from QP. 

\section{Solving QPs and LPs with distributed discrete dynamical systems}
\label{sec:theory}

Quadratic programming refers to 
\begin{align}
\label{eq:cost}
\textrm{minimize: } & f(x) = \frac{1}{2} x^T Q x + p^T x\ ,\\
\textrm{subject to: } & g(x) = Ax - k \le 0\,
\label{eq:constraints}
\end{align}

where $A\in \mathbb{R}^{M \times L}$, $x\in\mathbb{R}^L$, $p\in \mathbb{R}^L$,  $k\in \mathbb{R}^M$ and $Q\in \mathbb{R}^{L \times L}$. This also covers the sub-problem of linear programming (LP), wherein $Q = 0$. A quadratic program is convex when the feasible set is a convex set and $Q$ is symmetric positive semi-definite ($Q\in S^L_{+}$). 
In the following \hyperref[sec:iterative_solvers_overview]{section (\ref{sec:iterative_solvers_overview})}, we go over an iterative strategy to solve the QP from eqs.~\eqref{eq:cost} and~\eqref{eq:constraints}. In \hyperref[sec:implementation]{section \ref{sec:implementation}}, we present the implementation of our neuromorphic algorithm to solve convex QP such that it respects the constraints and strengths of Loihi 2, as determined in previous studies\cite{loihi2021survey}.

\subsection{Iterative solvers for convex quadratic programs}
\label{sec:iterative_solvers_overview}

\begin{figure*}
    \centering
    \includegraphics[width=\textwidth]{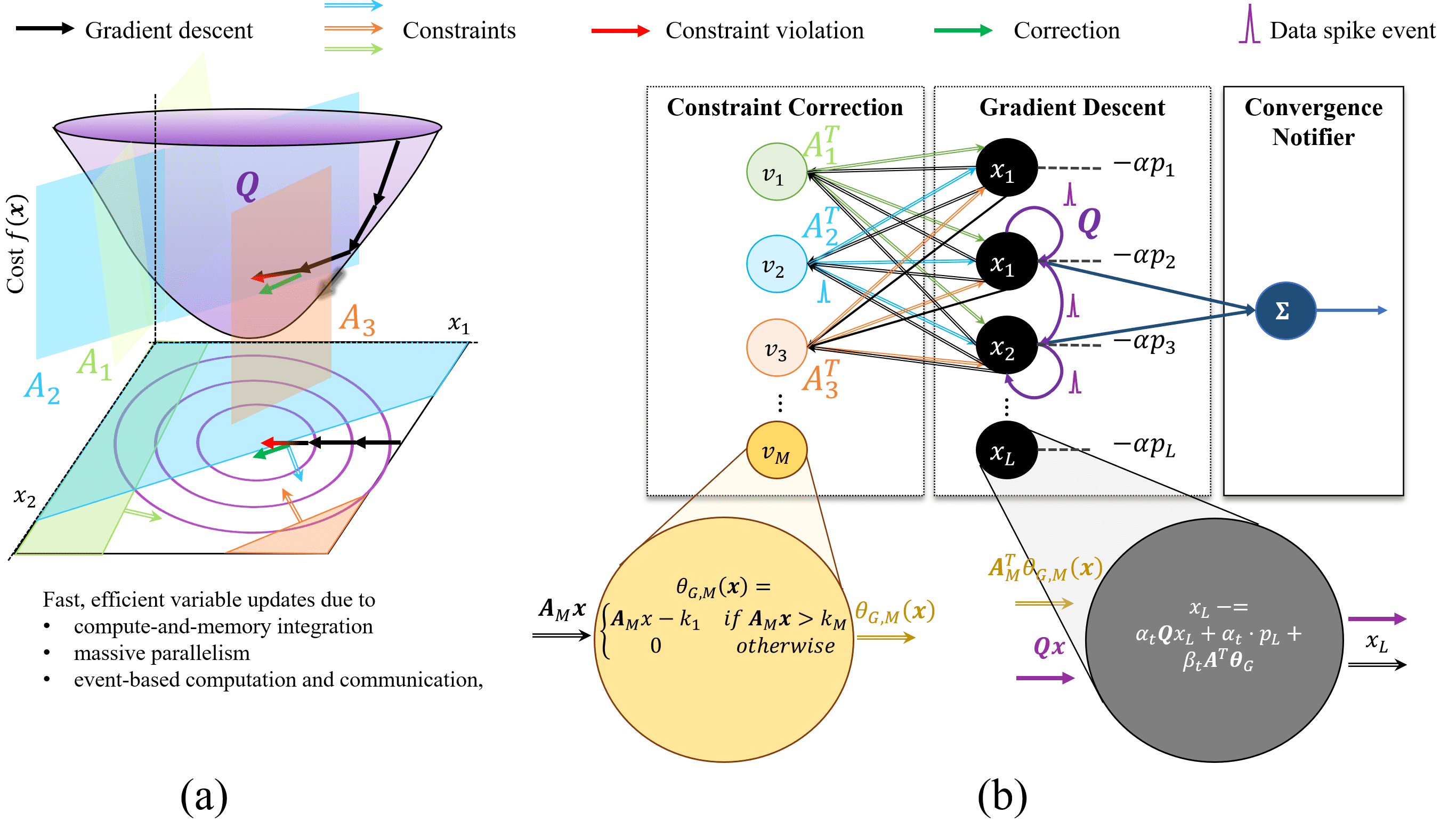}
    \caption{The QP solver can be formulated as a spiking neural network.
    a) represents the two-dimensional projection of a convex region defined by $Q$, 
    as contour lines, alongside constraint planes defined by $A$ as shaded regions. Black arrows represent the trajectory of gradient descent until it crosses the blue constraint plane, violating it by an amount proportional to the red arrow. At that point, gradient descent is corrected by the light blue arrow contribution which brings the network state back to the feasible region (final green arrow).
    b) The Gradient Descent block shows the network encoding of a solver for unconstrained QP. Each black circle represents a neuron that performs gradient descent updates, dashed lines represent biases and arrows synapses. For constrained QP, we add the Constraint Correction block, where colored circles are the neurons computing constraint violations and corresponding corrections for each constraining plane. The correcting terms are sent to the gradient descent neurons via channels indicated as colored arrows. An integration neuron (navy blue) tracks the cost of the current network state. In case of PIPG, each of the $L$ gradient descent neurons encodes one scalar entry $x^i_t$ of the variable vector $x_t$. It receives input $x_t$ from the other gradient neurons via synapses with multiplicative weights defined by the vector $Q_{i}$. In addition, it receives input $v_t$ from the constraint neurons via synapses of weight $A^T_i$. Its state is updated according to eq.\ \eqref{eq:pipg_x} and returned to the other gradient neurons. Each of the $M$ constraint neurons maintains internal states $w^j_t$ and $v^j_t$. It receives the states $x_{t}$ from all gradient descent neurons via synapses with multiplicative synaptic weights $A^T_j$. It then updates its internal states $w^j_t$ and $v^j_t$ according to eq.~\eqref{eq:pipgeq}-\eqref{eq:w_j}.}
    \label{fig:encoding}
\end{figure*}

The unconstrained QP defined by eq.~\eqref{eq:cost} alone can be solved by first-order gradient descent. The convexity of the unconstrained problem guarantees the convergence to the global minimum: 
\begin{equation}
x_{t+1} = x_t - \alpha \cdot \nabla f(x) = (I - \alpha Q)x_t - \alpha p\ .
\label{UQPGD}
\end{equation}
The constant $\alpha >0$ determines the step size of the gradient descent.

When constraints are introduced, as defined by eq.~\eqref{eq:constraints}, pure gradient descent according to eq.~\eqref{UQPGD} may lead to constraint violations in the course of iterations. Mathematically, a constraint violation occurs if a hyperplane $A_jx=k_j$ is crossed (the hyperplane is defined by the normal vector $A_j$, which is the $j^{th}$ row of $A$ \cite{mancoo2020convexspiking}). To avoid such violations, one can deflect the gradient descent dynamics into the direction of the normal vector if a constraint boundary is crossed. The dynamics are explained in fig.\ \ref{fig:encoding}a. 

To ensure that the gradient dynamics evolves in the feasible region, it is a sufficient condition to add this correction to our dynamical system,
\begin{align}
    \label{eq:dynamics}
    x_{t+1} &= (I -\alpha_t Q) x_t - \alpha_t \cdot p -\beta_t A^T \theta_G(x_t) \\
    \label{eq:gradient_constraints}
    \theta_G(x_t) &= 
    \begin{cases}
  A x_t - k  & \text{ if }  A x_t > k \\
  0 & \text{ otherwise.}
\end{cases}\ ,
\end{align}
where $\theta_G(x_t)$ denotes the relu-function 
indicating when $x_{t+1}$ crosses a constraint hyperplane during gradient descent. Here, the hyperparameter $\alpha_t$ decays while $\beta_t$ grows over time as the solver approaches the minimum of the state space.

The convergence rate can be accelerated by an additional integral term that accumulates the deviation of the state vector outside the feasible region, i.e., the constraint violation, similar to a proportional integral controller. With this additional feature, one can revise the dynamics as \cite{yue2020pipg},

\begin{align}
    \label{eq:pipg_x}
    x_{t+1} &= \pi_\mathbb{X} (x_t - \alpha_t(Q x_t + p + A^T v_t))\ ,\\
        \label{eq:pipgeq}
    v_{t} &=  \theta_G(v_{t-1}) (w_{t} + \beta_t(Ax_{t} - k))\ , \\ 
    \label{eq:w_j}
    w_{t+1} &= w_t + \beta_t( Ax_{t+1} - k)\,
\end{align}
where $\pi_\mathbb{X}$ refers to the projection of the corrected $x_t$ into the feasible set $\mathbb{X}$. The hyperparameters $\alpha_t$ and $\beta_t$ depend on the curvature of the cost function \cite{yue2020pipg}.

The dynamics described by eq.~\eqref{eq:pipg_x}-\eqref{eq:w_j} can be interpreted as that of a 2-layer event-based recurrent neural network as illustrated in fig. \ref{fig:encoding}b which minimizes the cost function \eqref{eq:cost} using gradient descent. After algebraically eliminating $w_t$, the variables $x_t$ and $v_t$ can be identified as state variables of different types of neurons, while $Q$ and $A$ can be identified as matrices representing (sparse) synaptic connections between those neurons. 
Relying only on element-wise or matrix-vector arithmetic, this type of network can be efficiently implemented on neuromorphic hardware.

\subsection{Neuromorphic hardware implementation}
\label{sec:implementation}

Neuromorphic hardware architectures are designed to efficiently execute event-based neural networks. Example systems include SpiNNaker 1 and 2, Dynaps, BrainScaleS 1 and 2, TrueNorth, and Loihi 1 and 2 \cite{ashresthtasurvey2022}. A shared characteristic of many of these architectures is a large number of parallel compute units operating out of local memory, executing either fixed-function or highly programmable neuron models. Typically, these architectures are optimized for sparse event-based computation and communication. 
The use of the Loihi 2 for solving QP problems in this study has been motivated by prior work, showing that the Loihi 2 architecture excels at solving iterative constraint optimization problems\cite{loihi2021survey}.

\paragraph{Mapping the QP solver to Loihi 2}

\begin{figure*}
    \centering
    \includegraphics[width=\textwidth]{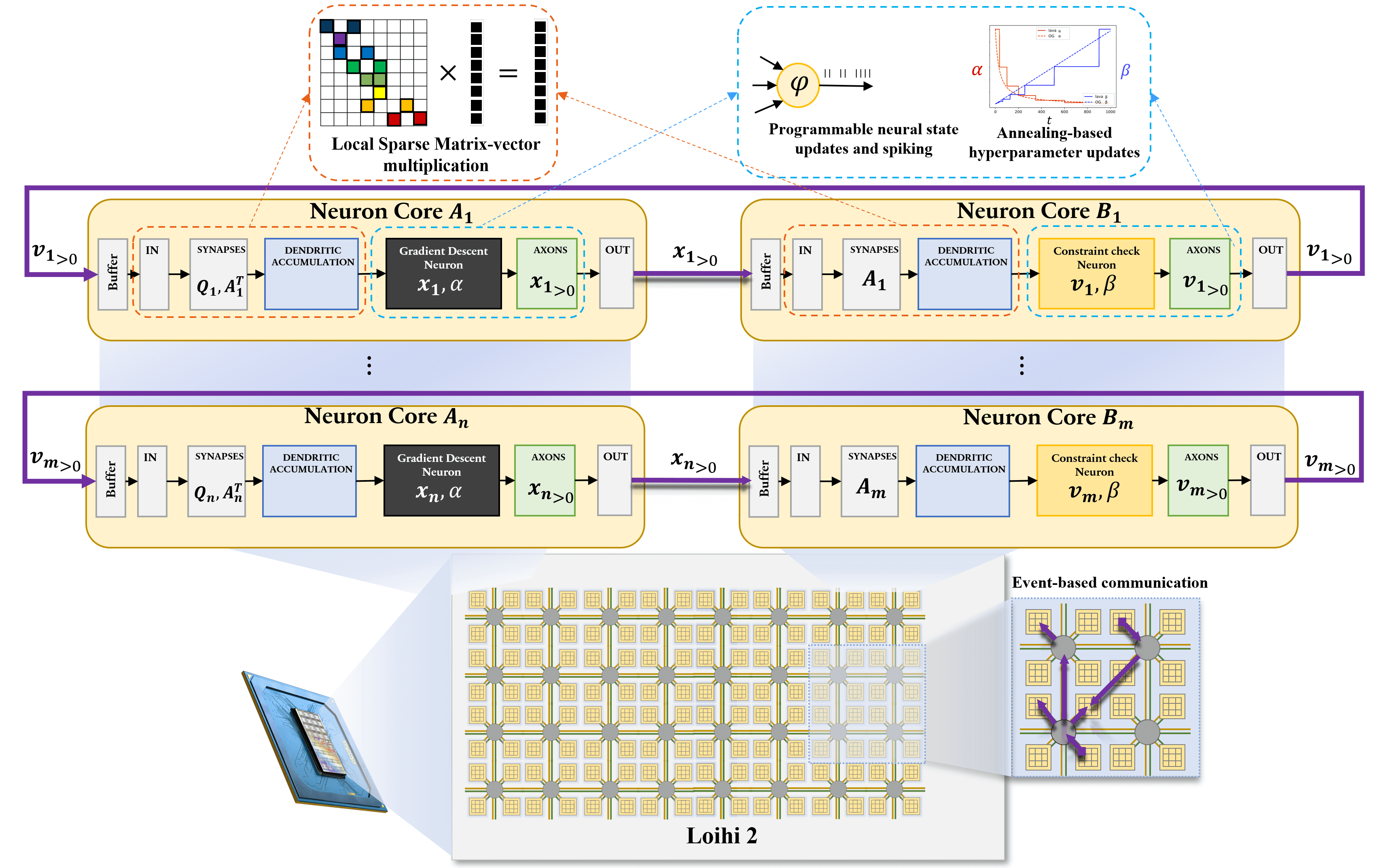}
    \caption{Mapping of the QP/LP solver to a neuromorphic substrate, an Intel Loihi 2 chip is shown here. Each Loihi 2 chip supports up to 1M neurons and up to 120M synapses depending on the model complexity, spread across 128 computing cores. Each neuron integrates synaptic-weighted input spikes from pre-synaptic neurons, can update local state variables via micro-code programmable operations, and can send integer valued message payloads to other post-synaptic neurons. Each core can house multiple neurons and all these groups of neurons (and the synaptic connections corresponding to them) are spread across multiple cores.  The recurrent dynamics  $x_{t+1}=h(x_t)$ described by Eq.~\eqref{eq:pipg_x}-\eqref{eq:w_j} combine gradient descent and constraint corrections and are mapped to the colorful and black neurons illustrated in \hyperref[fig:encoding]{Fig.~\ref{fig:encoding}b}, respectively. All of these neurons are updated in parallel by different cores on the chip, shown here as yellow blocks. Multiple cores $A_1$ to $A_n$ update the gradient descent neurons described by Eq.~\eqref{eq:pipg_x}, while multiple cores  $B_1$ to $B_m$ update the constraint correction neurons described by Eq.~\eqref{eq:pipgeq}-\eqref{eq:w_j}.
    The network state evolves towards the minimum in the energy landscape (Fig.~\ref{fig:encoding}a ) by exchanging spikes (bottom right zoom-in). The final solution is then post-processed to get the solution for the orignal non-preconditioned problem which is then used in the next stage of the ANYmal control pipeline.}
    \label{fig:implementation}
\end{figure*}

The QP solver corresponding to the dynamics of eq.~\eqref{eq:pipg_x}-\eqref{eq:w_j} was implemented on the second generation of the Intel Loihi research chip \cite{orchard2021} (\hyperref[fig:implementation]{fig.~\ref{fig:implementation}}). The massively parallel chip architecture consists $~$128 independent asynchronous cores that communicate with each other by exchanging up to 24-bit messages (graded spikes) using local on-chip routers. 
The innards of a core are shown in \hyperref[fig:implementation]{fig.~\ref{fig:implementation}}. Ingress spikes from other cores are first buffered before passing through the Synapse stage. This stage effectively performs a highly optimized dense or sparse matrix vector multiplication with up to 8-bit synaptic weights and up to 24-bit spike activation. The resulting product can be read by the Neuron stage from the Dendritic Accumulator stage. The Neuron stage executes stateful, parametrizable neural programs supporting basic arithmetic and bit-wise operations on variables up to 24 bits as well as conditional logic. Programs can also generate egress messages which are routed to other cores via the Axon stage.

Variables $x_t$ and $v_t$ defined in eq.~\eqref{eq:pipg_x}-\eqref{eq:w_j} are encoded as 24-bit state-variables of two different types of neurons in the chip, i.e., the gradient descent and constraint check neurons respectively.
As seen in fig. \ref{fig:implementation}, the spikes, in this case, carrying the values of states $x_t$ and $v_t$, are multiplied with the synaptic weights,  $Q$ and $A$,  and $A^T$ respectively the results of which are fed into the gradient descent and constraint check neurons respectively.  The neuron programs then perform the remaining arithmetic operations involved to complete the dynamics. This co-location of memory and compute in the neuromorphic chip leads to gains in terms of energy and time to solution for the algorithm. Note that different types of QP solvers can be implemented by merely changing the state update dynamics of the constraint correction or gradient descent neurons which is possible due to Loihi 2's programmable neurons. The updated states are only communicated to the next neuron through synapses if necessary, in this case, when the value is non-zero. 
 
 According to Yue et al. \cite{yue2020pipg} the hyperparameters, $\alpha$ and $\beta$, need to evolve at a certain rate for quicker convergence.This evolution, however, makes use of general-purpose division and floating-point operations which are not supported on Loihi 2. We emulate the evolution on Loihi 2 by halving $\alpha$ and doubling $\beta$ at a schedule, i.e, implementing simulated annealing for the hyperparameter $\alpha$ and geometric growth for $\beta$. The decay by halving is achieved using a bit right-shift operation which is the same as division by 2. The vector of values of the state variables of the gradient descent neurons after convergence of the dynamics of the network is the solution to the QP. Note that, in order to accelerate the convergence, QP problems are often first preconditioned. We choose the Ruiz preconditioner which is known to work well for block-diagonal problems like the ones we deal with this in this article. The pre-conditioning procedure is carried out on the host CPU in our experiments.

For QPs arising in MPC, the neuromorphic architecture of Loihi 2 already offers advantages over conventional architectures and ANN accelerators because of the compute and memory co-location which will be further discussed in section \ref{sec:mpc}. However, first-order type algorithms like those in eq.~\eqref{eq:dynamics}-\eqref{eq:w_j} can be made more suitable for neuromorphic hardware by employing $\Sigma$-$\Delta$ coding which can be implemented in Loihi 2 \cite{orchard2021efficient} . Doing so sparsifies the spiking activity at the cost of solution accuracy. This, in turn, improves the time and energy to solution because of fewer messages/spikes being moved around and, as a consequence, fewer operations in general. However, this reduces operations within acceptable solution degradation only when the datatypes on chip are more precise (16-/32-bit datatypes) which is currently not the case in Loihi 2. Therefore, in this article we focus on results obtained without employing sigma-delta coding.

\paragraph{Programming Loihi 2 with Lava} 
The QP solver is implemented on Loihi 2 using the open-source software framework \textit{Lava} for neuromorphic computing in the lava-optimization library. At its core, programming in Lava is based on asynchronous processes that communicate with each other via message-passing over channels. Lava provides a cross-platform runtime and compiler to execute algorithms on different backends such as CPU/GPU and Loihi 2, but is also open to extension to other neuromorphic platforms. Lava-optimization is one of several high-level algorithm libraries that build on top of Lava.\footnote{Lava is an open-source software licensed under BSD 3-Clause and LGPL 2.1, and only proprietary modules required to run code on Loihi 2 are confidential to members of \textit{Intel's Neuromorphic Research Community} (\textit{INRC}). The proprietary code required to execute the solver with high performance on Loihi 2 can be accessed after joining the INRC.
}

.

\section{Quadratic programming for model predictive control}
\label{sec:mpc}
To demonstrate the value of our neuromorphic QP solver,  we apply it to solving the QPs arising in model predictive control of a state-of-the-art quadrupedal ANYmal robot \cite{hutter2016anymal}. 

\subsection{The ANYmal robot}
The ANYmal platform comprises a series of quadrupedal robots designed for different kinds of tasks like inspection, surveillance, and search and rescue missions. The robots are equipped with a suite of sensors enabling autonomous navigation and perception of their surroundings, as well as a comprehensive software ecosystem. The robot is commercially available and deployed in a range of industrial and commercial settings. However, model predictive control for this robot has on the order of $10^3$ variables. Due to this high dimensionality, each robot requires a separate Intel\textregistered
~Core\texttrademark~i7 processor in order to solve model predictive control within acceptable time budgets. Further, as a mobile platform, bringing down power utilization would contribute to longer uptimes. Loihi could be used as a co-processor to solve model predictive control with very little additional burden on the battery. We therefore use data acquired from a physical ANYmal robot performing tasks provided by the ANYmal team.

\subsection{Problem definition}
At the core of an MPC iteration is the mathematical optimization problem of minimizing the error between actual and goal trajectories, i.e., the tracking error. The optimization is subject to constraints, such as the limits posed by the rigid-body dynamics, joints, actuators, and environment. As such, the optimization problem is a computationally expensive, nonlinear (and perhaps non-convex). However, it is typically `linearized-and-quadratized' using techniques akin to Taylor series expansion, resulting in a convex QP problem from eq.~\eqref{eq:cost} and \eqref{eq:constraints}. MPC is a computationally intensive control scheme that must be executed under strict time budgets in real-time control loops. Solving the QP accounts for a major chunk of computational time in MPC of ANYmal. The QP further increases in complexity with larger time horizons covered by the MPC or more degrees of freedom of the robot. In order to meet these requirements, MPC is solved on a dedicated laptop-class CPU. Running the QP in the MPC on Loihi would be the first step towards making computation for control more suitable for SWaP systems.  

MPC uses a mathematical model of ANYmal to predict the temporal evolution of its state over a time horizon in the future. The prediction is based on the optimization of a task-specific objective function and the robot's current measured state (its position, orientation, linear and angular momenta, and joint angles \cite{sleiman2021unified}). From the predicted time-series of the control variables, only a fraction in the beginning is used for actuating the locomotion of ANYmal. For example, a typical MPC horizon in the control loop of ANYmal of about 1 second is split into 100 time-steps of 10 ms each and only first 20-30 ms worth of predictions are used to issue control commands. Once the robot moves, a new state measurement and sensory data are fed back to the model to generate the next set of predictions, thus completing an MPC iteration.

\subsection{Neural and synaptic scaling}

For an MPC horizon of $N=100$, we require $7248$ neurons connected through $405,504$ weights at most. These resources are well within a single Loihi 2 chip's capacity of at least 64kB of memory synaptic connections and 8kB for neural programs per neuron core. Assuming 8-byte neurons and 1-byte weights, we get $\sim$$1$ million neurons and $\sim$$15$ million 8-bit synapses for a single chip with $\sim$$123$ cores. Note that the complexity of the neuron program can lead to higher memory requirements for it. For this article, we implemented programs that require 2 bytes per neuron resulting in 500,000 available neurons in a single chip. This is well above the requirements for the QPs covered here. \par
To understand the scaling of neurons and synapses on Loihi for our QP formulation, we first calculate the number variables and weights involved in the QP problem for MPC. ANYmal's locomotion is captured by the temporal evolution of a 48 dimensional {\em attribute vector} formed by concatenating a 24 dimensional control vector (12 joint velocities and 12 contact forces) and a 24 dimensional state vector (6 base poses, 6 momenta, 12 joint angles). When we formulate the QP problem for an MPC iteration of horizon with $N$ time-steps, we treat all $N$ values of the attribute vector at every time-step as a flattened vector. The causal dependence between the attribute vectors at successive time-steps is captured in constructing the cost, $Q$, and constraints, $A$, matrices of the QP problem from eq.\ \eqref{eq:cost} and \eqref{eq:constraints}. We need one neuron for each decision variable and one for each constraint variable. Therefore, the number of neurons required to map QPs of this type onto Loihi is given by:
\begin{equation}
    n_{neurons} = (N+1)* n_{states} + N*n_{controls}, 
    \label{eq:num_neurons}
\end{equation}
where $n_{states}$ is the size of the state vector (24 for ANYmal) and $n_{control}$ is the size of the control vector (24 for ANYmal). Further, the maximum number of synapses required to map this QP onto Loihi is given by,

\begin{equation}
    \begin{aligned}
            n_{synapses} = n_{states}*(2*n_{states} + n_{controls})*N \\ + (n_{states} + n_{controls})^2 * N + n_{states}^2.
    \end{aligned}
    \label{eq:num_weights}
\end{equation}

\begin{figure*}
    \centering
    \includegraphics[width=\textwidth]{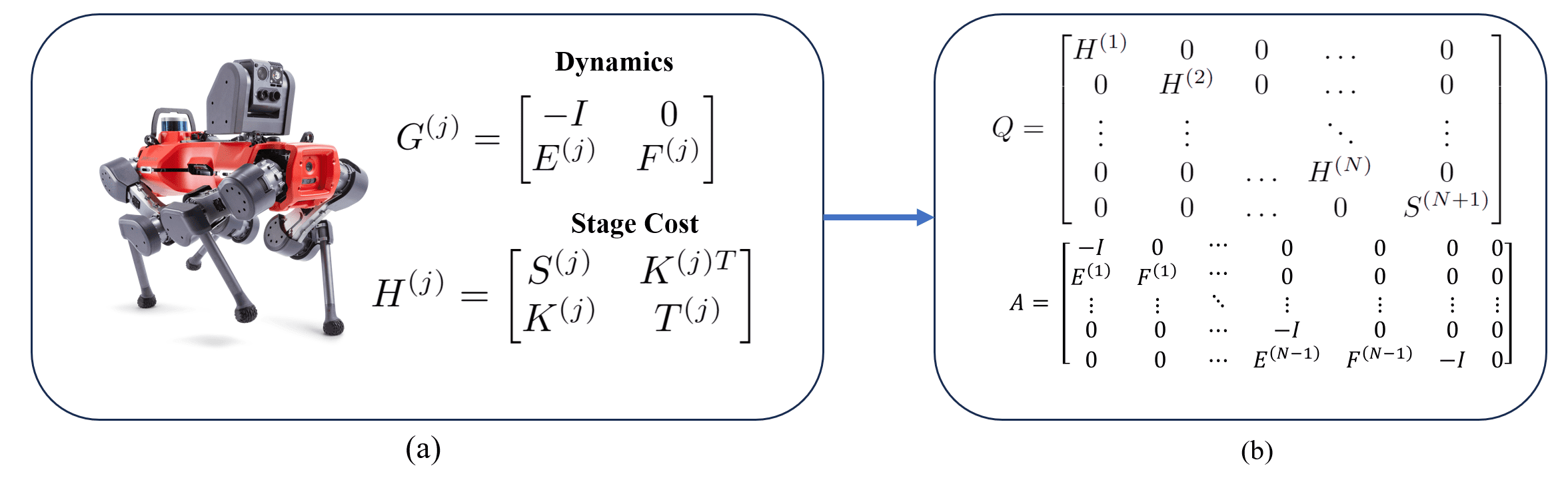}
    \caption{ In (a) the matrices pertaining to stage costs, $H^{(j)}$,  and dynamics , $G^{(j)}$, of ANYmal are associated with every timestep of the MPC. $E^{(j)}$ and $F^{(j)}$ represent the dynamics of the robot at timestep $j$. $S^{(j)}$,  $K^{(j)}$ and $T^{(j)}$ are the stage costs at timestep $j$. It is seen in (b) how these matrices are concatenated together to form large $Q$ and $A$ matrices for cost and constraints of the QP (Eq. \ref{eq:cost} \& \ref{eq:constraints}) and respectively. Note that the matrices in (b) are sparsely populated making these types of problems well-suited for Loihi.}
    \label{fig:matrices}
\end{figure*}

\subsection{Benchmarking procedure and dataset} 
\begin{figure*}
    \centering
    \includegraphics[scale=0.13]{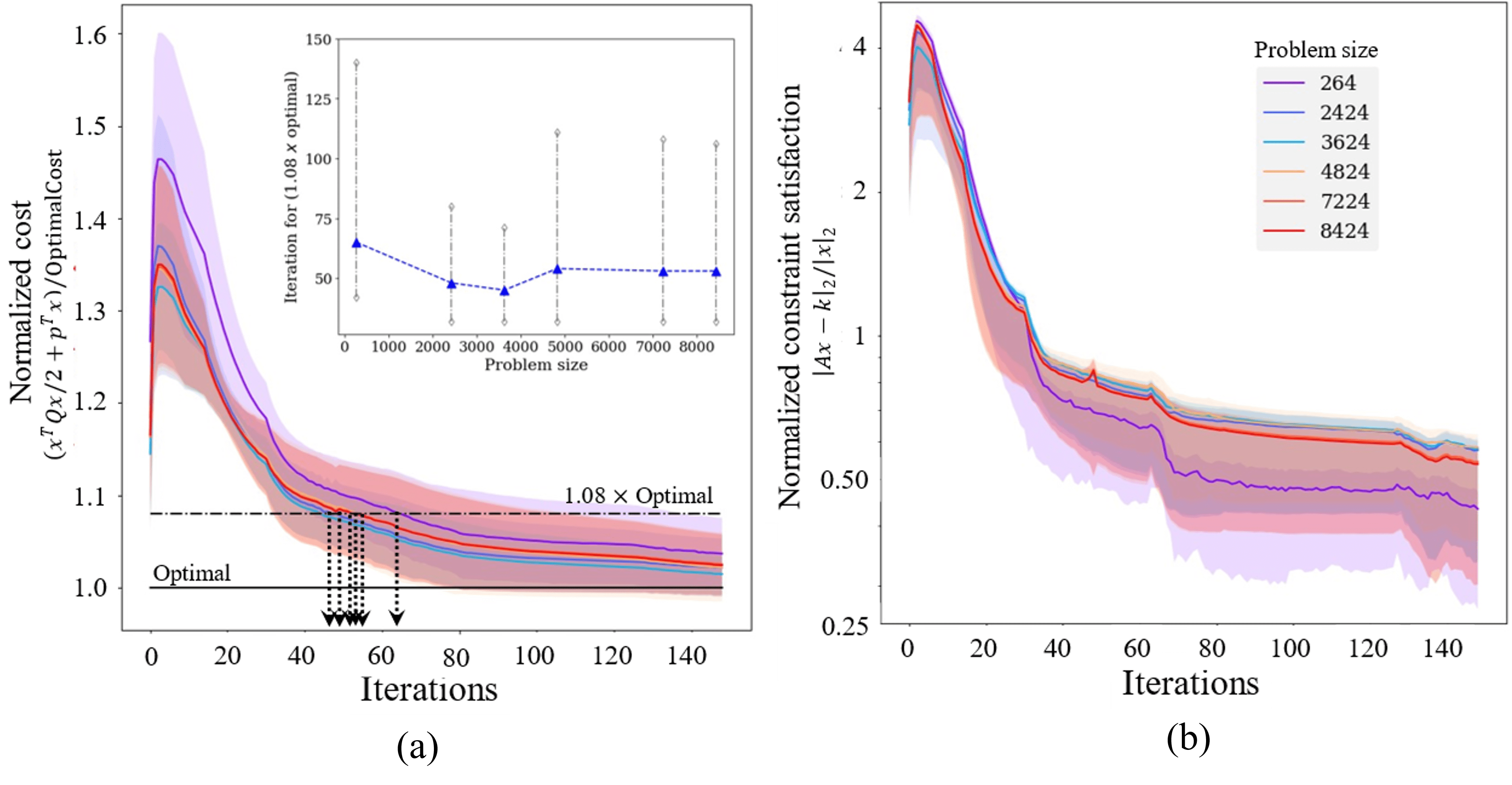}
    \caption{Convergence analysis of Lava QP solver for 60 QPs in the MPC loops of ANYmal as measured by (a) the cost eq.~\ref{eq:cost} normalized by OSQP's optimal cost and (b) number of constraint violations eq.~\ref{eq:constraints} normalized by the L2-norm of the final solution is shown here.
    Solid lines represent the mean for cost and constraints respectively for 60 different problem sizes $N = {5, 50, 75, 100, 150, 175}$ translating into 264 to 8424 QP variables as indicated by the legend. Shaded regions represent one standard deviation from the mean. The horizontal black lines corresponds to the optimal cost.
    The inset in (a) depicts the number iterations required for different problem sizes to reach an optimality gap of 8\% with respect to OSQP.}
    \label{fig:cost_csat}
\end{figure*}
The efficiency, scalability, and speed of the Loihi QP solver was benchmarked against 1) CVXOPT, 2) SCS solver running on a laptop CPU and OSQP running on  3) a standard CPU,  4) an embedded CPU, and 5) against the CUDA-optimized version of it, cuOSQP, running on a GPU. \footnote{Loihi 2: Oheo Gulch board running Lava v0.7.0 and Lava-Optimization v0.2.4 with an Intel Core i7-9700K CPU host with 32GB RAM running Ubuntu 20.04.5 LTS. \newline OSQP: OSQP v0.6.2.post9 ran on an Intel Core i7-9700K CPU @ 3.6GHz with 32GB DRAM running Ubuntu 20.04.5 LTS and an Nvidia Jetson Orin 6-core Arm Cortex-A78AE v8.2 64-bit CPU
1.5MB L2 + 4MB L3 CPU with 6GB of shared RAM running Ubuntu 20.04.6 LTS. \newline
cuOSQP: ran on an Intel Core i7-9700K CPU @ 3.6GHz with 32GB DRAM running Ubuntu 20.04.5 LTS with an Nvidia GeForce RTX 2070 Super GPU with 8 GB of RAM and cuda 10.2.
}
Evaluation metrics were  time to solution (TTS), energy to solution (ETS) and energy-delay product (EDP) for different problem sizes. OSQP is a CPU-based state-of-the-art solver for convex QP problems with linear inequality constraints \cite{stellato2020}. We ultimately use the OSQP solver as a reference for our Loihi-based solver since it was the best performing CPU solver.

The dataset used in this paper consists of data from 2173 individual time-steps of MPC of an ANYmal robot \cite{OCS2}. The dataset consists of the matrices obtained after linearization and quadratization of a non-linear objective function and penalizes deviation from a reference trajectory as well as the ANYmal dynamics. We tile these matrices in an appropriate manner to construct the $Q$ matrix of the QP objective function from \eqref{eq:cost} and the $A$ matrix of the constraints from \eqref{eq:constraints}. This tiling is demonstrated in fig.~\ref{fig:matrices}. The derivation and explanation of this tiling has been omitted here since it's beyond the scope of the article. It can be seen that these matrices are mostly block-diagonally populated and very sparse. This entails sparse matrix-vector multiplication making a suitable candidate for Loihi 2. The vectors $p$ and $k$ from the cost and constraints respectively are constructed by stacking the vectors associated with the tracking cost and the dynamics in each stage of the MPC. We investigate how well the performance of the solver scales with problem size by varying the number of variables. In general, the problem size is determined by the time horizon and the degrees of freedom of the robot. Here, we choose six horizon lengths $N = {5,\, 50,\, 75,\, 100,\, 150,\, 175}$ time-steps, resulting in problems with $264,\, 2424,\, 3624,\, 4824,\, 7224,\, 8424$ variables, respectively. We use this as a proxy for a robot with higher degrees of freedom.  Most QPs in the dataset are similar in difficulty (similar condition numbers).  Therefore, for each horizon length, we have chosen 10 different QPs of varying difficulty that best represent all the problems in the dataset. In summary, we have a dataset containing a total of 60 representative QP problems of 6 different sizes from the ANYmal dataset, spanning from $\sim250$ to $\sim8500$ variables. 

\subsection{Convergence, power and performance analysis}


In the context of solving a type of QP problem, i.e, LASSO problems, on Loihi \cite{davies2018loihi, loihi2021survey} Loihi-based solvers rapidly converge to approximate solutions, but limited precision available on the chip, explained in \ref{sec:implementation}, limits the convergence to a finite optimality gap. The same applies to Loihi 2 as well. However, the CPU solver operates with 64-bit floating-point precision.To make a fair comparison across all platforms, we run all solvers until they have converged up to the same accuracy of 8\% from the true solution. This accuracy was chosen in alignment with the precision needed to make MPC robust for various control applications \cite{diehl2007stabilizing}. For MPC applications that need higher accuracy, we can in general increase the precision of the Loihi 2 solution. For this, Loihi 2 could represent each 64-bit variable by allocating several of its 24-bit states, and each 64-bit matrix weight by combining several of its 8-bit synaptic weights. To verify that successive MPC iterations converge stably even with limited precision, we randomly perturbed $Q$ and $A$ matrices for each successive iteration and verified that the cost and constraint satisfaction keep converging with warm-starting of the QP solver. This suggests that approximate solutions are sufficient in closed-loop control applications. 

Fig.~\ref{fig:cost_csat} demonstrates that the Loihi QP solver rapidly converges within $\sim$55 iterations on average, where a solution is considered to have converged when the solver reaches within 8\% or less of the OSQP reference solver. The solver continues to converge further, closer to optimality, but with a slower convergence rate. Similar to the cost, constraint satisfaction continues to improve beyond our definition of convergence. In the subsequent performance comparison results, we have used the fact that Loihi QP solver converges at 55 iterations, and compared its performance with that of OSQP.

\begin{figure*}
    \centering
    \includegraphics[width=0.93\textwidth]{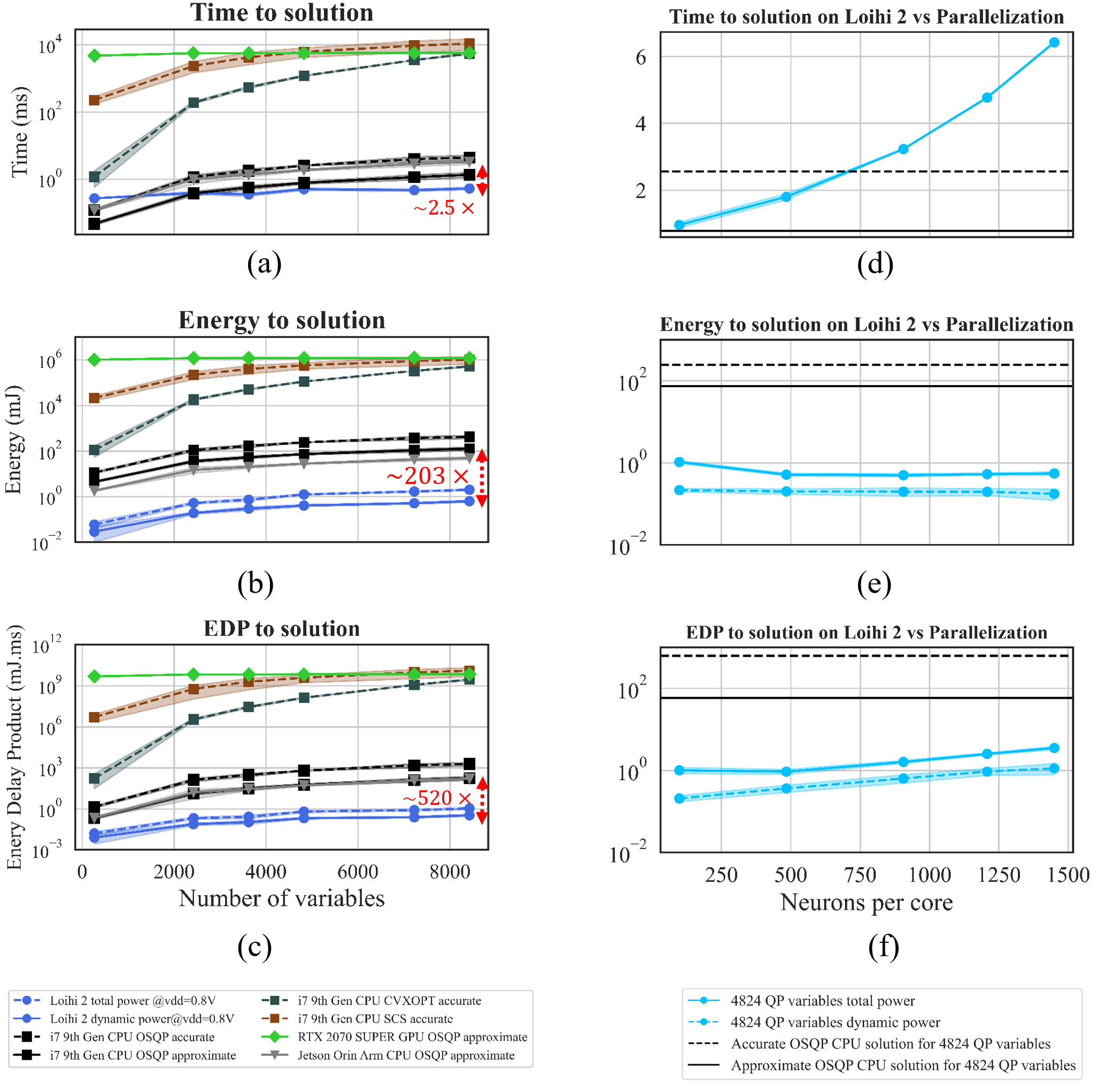}
    \caption{(a)-(c) demonstrate the advantages of the Loihi 2 based solver over the CPU solver on a laptop-class (black, dark slategrey, and brown) and edge-level CPU (gray) and also over its GPU-based implementation (lime green) for increasing problems size on different metrics. OSQP was the best performing solver and was therefore chosen as the reference. As seen from (d)-(f), there exists a parallelization-solution time trade-off but a slow solver beyond a certain point, could lead to increased EDP. Note that the speed of execution can be increased by operating at higher voltages. }
    \label{fig:T_E_EDP}
\end{figure*}

Fig.~\ref{fig:T_E_EDP}a shows the time-to-solution (TTS) for the same set of QP problems of increasing problem size. As the MPC time horizon $N$---and thus the number of QP variables---increases, TTS increases roughly linearly for both OSQP and the Loihi QP solver. While the OSQP solver solves small problems faster than Loihi 2 in the sub-millisecond regime, the time it takes OSQP to find the optimal solution (black dashed line) grows rapidly beyond Loihi's TTS. 
Let us consider, approximate solutions from Loihi for usage in iterative MPC with warm-starts. In this case, OSQP's TTS grows less rapidly but still exceeds Loihi towards the largest problem size. We see that the solution time for a standard GPU is orders of magnitude higher than even a CPU for problems of this scale for reasons mentioned in section \ref{sec:intro}. For the embedded CPU, the TTS follows the same trend as the Laptop CPU albeit even slower. 

Fig.~\ref{fig:T_E_EDP}b shows the corresponding ETS for increasing problem complexity. ETS for both OSQP and the Loihi QP increase with problem size as more and more cores on Loihi are utilized. Loihi achieves a significant advantage compared to laptop class CPUs in energy  of over $200\times$ for all problem sizes. Energy consumption of the Loihi QP solver is composed of static and dynamic energy: $E=E_{static}+E_{dyn}$. Static energy is mostly governed by chip leakage power on Loihi $E_{static}=P_{static}\cdot TTS \cdot c_{active}/c_{total}$ where $c$ is the number of active and total cores respectively. Measurements for this analysis have been obtained from early silicon samples that still have highly variable leakage characteristics leading to a high $P_{static}=1.4W$ per chip at the lowest operating voltage and thus is less relevant for this analysis. Dynamic energy is governed by the amount of operations within Loihi cores to update neurons and route traffic between cores. 

Further, fig.~\ref{fig:T_E_EDP}c shows a $520\times$ energy-delay product advantage for the largest problem size when using Loihi 2 in place of a laptop CPU. It can be seen in fig. \ref{fig:T_E_EDP}c that the EDP curves for CPUs (embedded and laptop) closely resemble each other. The laptop and embedded CPU thus trade TTS against ETS, but the combined EDP is characteristic for CPUs. For a GPU, both the ETS and EDP are orders of magnitude higher mainly.


By increasing Loihi's operating voltage from $VDD=0.6V$ to $0.8V$, Loihi's TTS improves by about 47\%. At the same time, dynamic energy increases by a factor of $1.9 \times$ for increasing operating voltage for a problem of 4824 variables. The total energy remains largely unchanged since the dominating leakage energy drops with lower TTS in this example.
\mbox{Fig.~\ref{fig:T_E_EDP}d-e} further illustrate how the level of multi-core parallelism allows to trade performance (TTS) for chip resources. As the number of neurons per core decreases and number of cores increases proportionately, TTS improves from 17.5 ms to 2.6 ms  for a problem size of 4824 variables. The slightly less than linear decrease in TTS results from additional multi-core synchronization and data traffic between parallel Loihi cores. Loihi's highly configurable nature allows users to trade off energy consumption (ETS), speed (TTS) and chip resource utilization, which can be useful in size-, weight- and power- (SWaP) constrained applications to select the optimal task-specific operating point.

\section{Conclusion}

We have developed an efficient and scalable approach for solving convex quadratic constraint optimization problems with neuromorphic chips. The benefits of neuromorphic systems can be exploited for optimization algorithms by interpreting the structure of many iterative optimization approaches---such as gradient descent and primal-dual updates---as the dynamics of recurrent neural networks. 
To demonstrate the benefits of this framework, we implemented a QP solver on Intel's Loihi 2 research chip and made it publicly available as part of the Lava software framework. To illustrate its benefits, we applied the QP solver to tackle workloads that arose in the model predictive control pipeline of ANYmal during its normal operation. Our Loihi solver solved the workloads with similar speed as the state-of-the-art solver OSQP on a modern CPU \cite{de2019trajectory}, which was the best von-Neumann-based-candidate for comparison, for approximate solutions while consuming about two orders of magnitude less energy.
Most notably, time and energy to solution on Loihi scale better with increasing problem complexity than OSQP on CPU.

The performance, energy and scaling advantages on Loihi result primarily from its massively parallel, event-based, memory-integrated hardware architecture. Multi-core parallelism allows to quickly and independently update a large number of variables. The hardware is further optimized for arithmetic on sparse matrices, as present in most large real-world QP workloads \cite{cheshmi2020}. Its event-based computation and communication further eliminates redundant data traffic routing and computation. Together, these features support scaling to large problems sizes with less overhead than on conventional computer architectures. In addition, the memory-integrated compute architecture minimizes the energy and latency required to access data for algorithmic iterations. With the observed gains in computational efficiency, interpretable model-based controllers could see increased adoption over more opaque model-free controllers like reinforcement learning, thus increasing the safety of applications.

The most significant limitation of the current approach is the limited bit-precision of state variables and synaptic connections on neuromorphic architectures like Loihi. This can lead to lower solution optimality than solvers on conventional floating-point architectures. Nevertheless, our solver consistently achieved solutions that deviated by less than $8\%$ from the true optimal solution for QPs extracted during a real-world operation of a physical ANYmal robot. For applications that require a higher precision, the limited precision of the Loihi solver can be circumvented by allocating multiple low-bit variables to effectively achieve higher precision arithmetic or enabling higher precision arithmetic in general in future chip generations. Nonetheless, preliminary observations in closed-loop control simulations of ANYmal hint that fast approximate solutions are sufficient for iterative scenarios like MPC with sequential warm starts. We hypothesize this is partly because the result of each MPC cycle is only used to determine motor commands for the next few time steps, despite optimizing over time horizons of more than 100 time steps. After controlling the next few steps, new sensory recordings are taken and the next MPC cycle can iteratively correct any errors resulting from, e.g., the linearized mechanical robotic model, environmental perturbations, and limited bit-precision. We invite robotics labs interested in assessing the performance of our fixed-point QP solver on their cyber-physical systems to become part of Intel’s Neuromorphic Research Community (INRC). By joining the INRC, labs will gain access to the proprietary code for running the solver on Loihi 2. They can then apply for borrowing a Loihi 2 board for closed-loop testing in their MPC applications. The implementation of our solver in Lava, a software framework built to support asynchronous message passing, bodes well for integration with the asynchronous publish and subscribe system of ROS. Further, with floating-point support, the current gains observed with compute-memory co-location can be bolstered by employing sigma-delta coding which would reduce spiking activity in the chip and consequently TTS and ETS.\par 

In applications where faster solves are required, the massive energy advantage of neuromorphic architectures can be traded for additional speed by executing multiple solver instances in parallel with different initial conditions and selecting the best and fastest solution. In general, the customizability of neuromorphic architectures like Intel's Loihi 2 allows to trade solution optimality, energy, speed, and chip resource utilization seamlessly against each other to achieve optimal operating conditions depending on the requirements of the task.

\section*{Acknowledgments}
We would like to thank Farbod Farshidian, a former member of the Robotics Systems Lab, ETH Zürich, for providing the data used in this article and for his inputs on ANYmal. We appreciate the valuable feedback on the manuscript from Prof. Alin Albu-Sch\"{a}ffer from the German Aerospace Center (DLR) and TU Munich. We thank Yulia Sandamirskaya (ZHAW, Zurich) and  Akshit Saradagi (Luleå  University of Technology) who participated in disucssions and also provided some useful suggestions for our work. Sumit Shrestha Bam and Leobardo E Campos Macias from the Neuromorphic Computing Lab, Intel Labs provided support with Loihi and Orin platforms. We also thank all the  members of the Neuromorphic Computing Lab, Intel Labs in general for their support in developing features for Lava and for assistance with hardware-related issues.


\bibliographystyle{IEEEtran}
\bibliography{biblio}
\pagebreak

\begin{IEEEbiography}[{\includegraphics[width=1in,height=1.25in,clip,keepaspectratio]{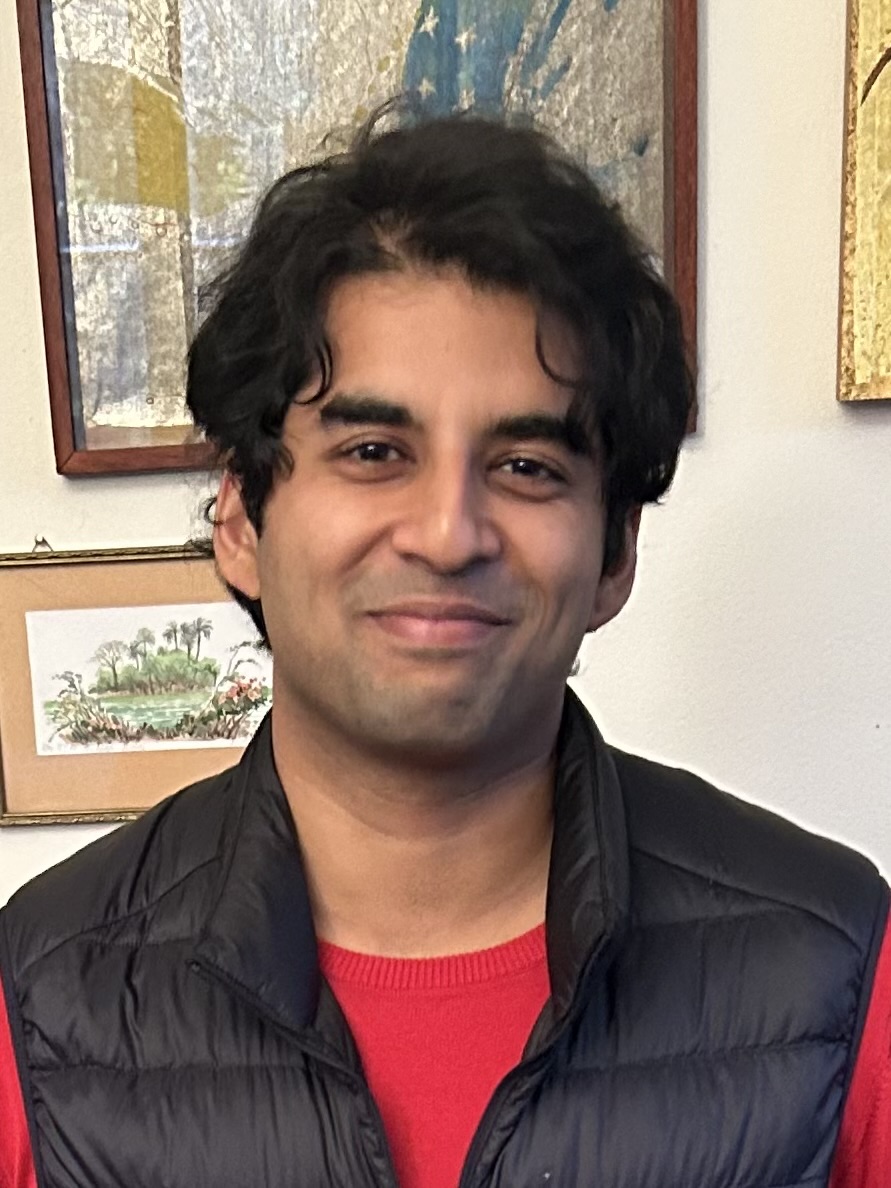}}]{Ashish Rao Mangalore}
completed his MSc in Neuroengineering at the the Technical University of Munich in 2022. He is currently a Doctoral Researcher at Intel Labs, Intel Corporation, Munich, Germany, working on algorithms for robotic control and planning with neuromorphic architectures. 
\end{IEEEbiography}

\vspace{11pt}

\begin{IEEEbiography}[{\includegraphics[width=1in,height=1.25in,clip,keepaspectratio]{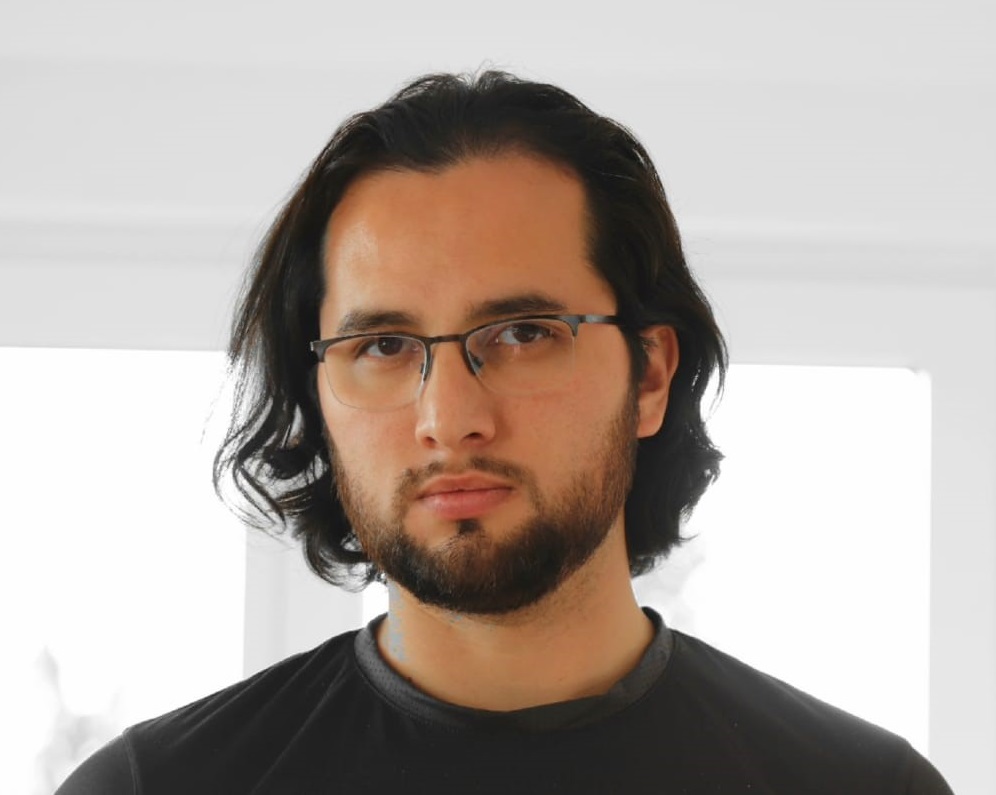}}]{Gabriel Andres Fonseca Guerra} received the Ph.D. degree in computer science with a focus on stochastic processes for neuromorphic hardware from The University of Manchester, Manchester, U.K., in 2020. 
He is currently a Research Scientist with Intel Labs, Intel Deutschland GmbH, Munich, Germany, working on algorithms and architecture development.

\end{IEEEbiography}

\vspace{11pt}

\begin{IEEEbiography}[{\includegraphics[width=1in,height=1.25in,clip,keepaspectratio]{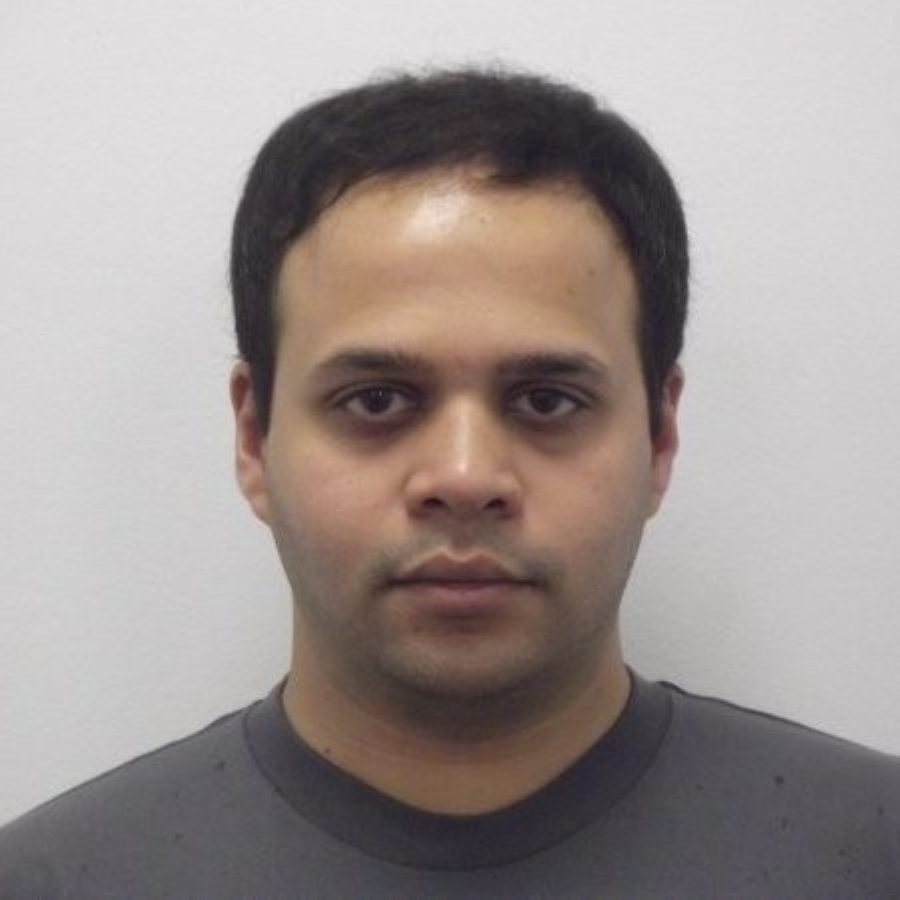}}]{Sumedh R. Risbud} received the Ph.D. degree in chemical and biomolecular engineering with a focus on theoretical fluid mechanics and microfluidics from Johns Hopkins University, Baltimore, MD, USA, in 2013.
He is currently a Research Scientist with Intel Labs, Intel Corporation, Santa Clara, CA, USA. His focus at the Neuromorphic Computing Lab is on algorithms and architecture development.

\end{IEEEbiography}

\vspace{11pt}

\begin{IEEEbiography}[{\includegraphics[width=1in,height=1.25in,clip,keepaspectratio]{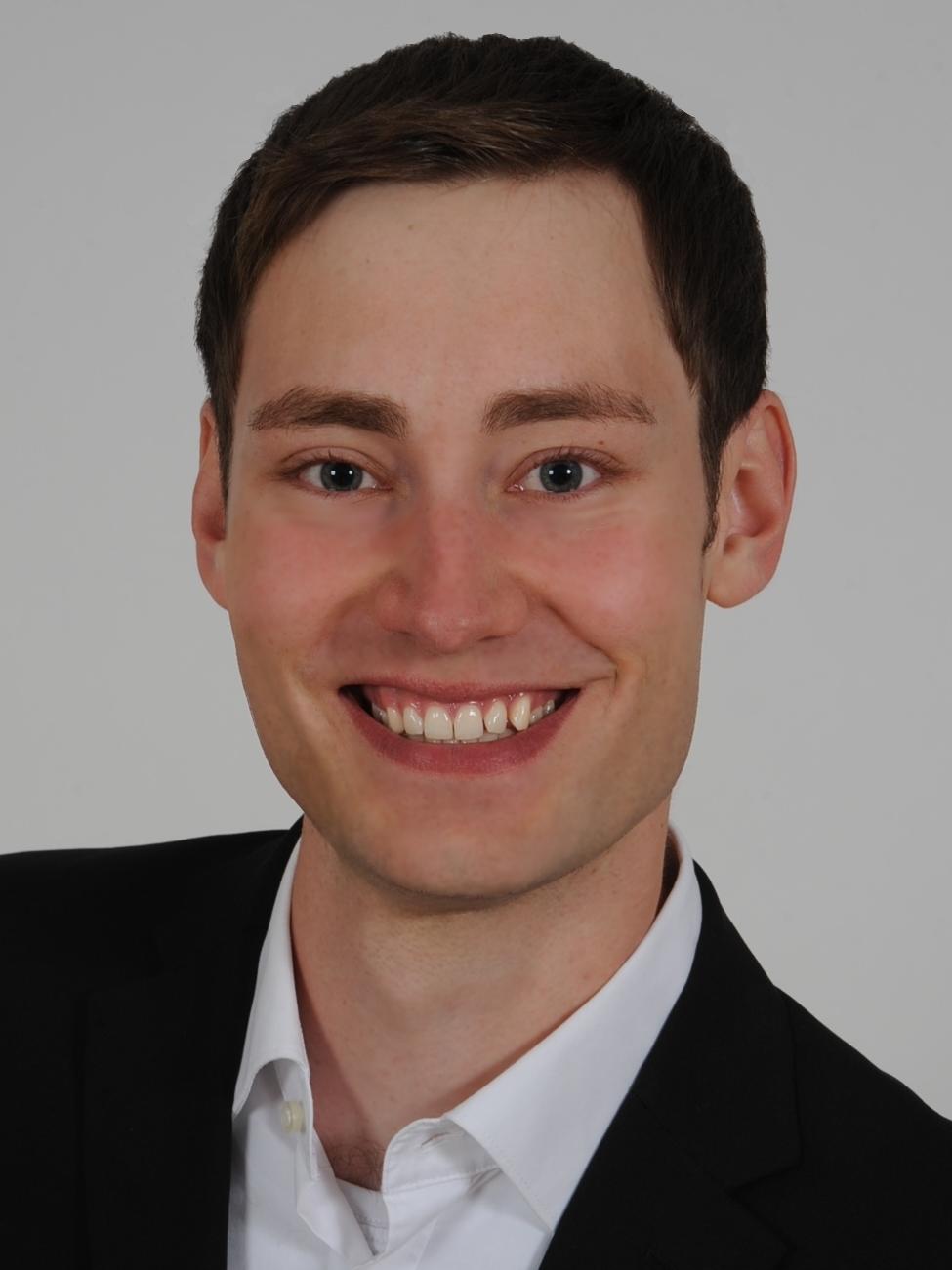}}]{Philipp Stratmann}
completed a Dr. rer. nat. in computer science at the the Technical University of Munich and the German Aerospace Center (DLR) in 2020. He is currently a Research Scientist with Intel Labs, Intel Corporation, Munich, Germany, working on algorithms and architecture modeling as a member of the Neuromorphic Computing Lab.
\end{IEEEbiography}

\vspace{11pt}

\begin{IEEEbiography}[{\includegraphics[width=1in,height=1.25in,clip,keepaspectratio]{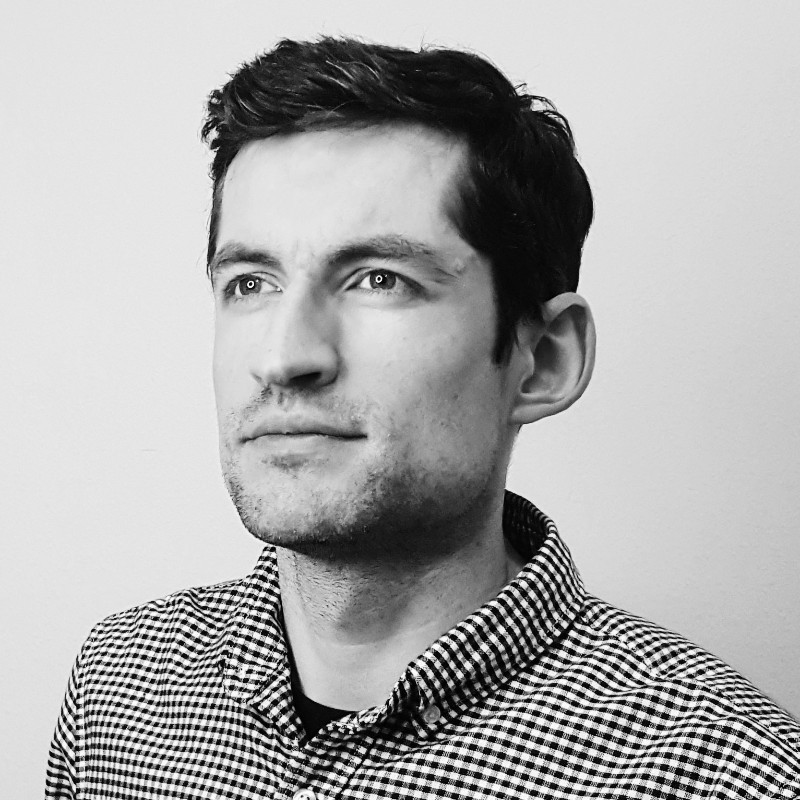}}]{Andreas Wild} received a Dr. rer. nat. degree in physics for the development of silicon-based electron spin qubits from the Technical University of Munich, Germany, in 2013.
Since 2015, he has been a Senior Researcher with the Neuromorphic Computing Lab, Intel Corporation, Portland, OR, USA, where he leads algorithm development. 

\end{IEEEbiography}

\vfill

\end{document}